\documentclass[letterpaper, 10 pt, conference]{ieeeconf}  

\IEEEoverridecommandlockouts                              

\usepackage{graphicx}
\usepackage{multirow}
\usepackage{subcaption}
\usepackage{acro}
\usepackage{url}
\usepackage{siunitx}
\usepackage{balance}
\usepackage{hyperref}

\usepackage{blindtext}

\title{\LARGE \bf
Reinforcement Learning for Blind Stair Climbing with Legged and Wheeled-Legged Robots
}

\author{Simon Chamorro$^{1,2}$, Victor Klemm$^{3}$, Miguel de la Iglesia Valls$^{4}$, Christopher Pal$^{1}$, and Roland Siegwart$^{2}$
\thanks{$^{1}$Department of Computer and Software Engineering, École Polytechnique de Montréal, 2900 Boul. Édouard-Montpetit, Québec, Canada. e-mail: {\tt\small simon.chamorro@polymtl.ca}}%
\thanks{$^{2}$Autonomous Systems Lab, ETH Zurich, Switzerland. }
\thanks{$^{3}$Robotic Systems Lab, ETH Zurich, Switzerland. }
\thanks{$^{4}$Ascento Robotics, Zurich, Switzerland. }
\thanks{Supplementary video: \href{https://youtu.be/Ec6ar8BVJh4}{https://youtu.be/Ec6ar8BVJh4}.}}

\DeclareAcronym{RL}{short=RL, long=Reinforcement Learning}
\DeclareAcronym{SLAM}{short=SLAM, long=Simultaneous Localization and Mapping}
\DeclareAcronym{GPS}{short=GPS, long=Global Positioning System}
\DeclareAcronym{MPC}{short=MPC, long=Model Predictive Control}
\begin{document}

\maketitle
\thispagestyle{empty}
\pagestyle{empty}
\begin{abstract}

In recent years, legged and wheeled-legged robots have gained prominence for tasks in environments predominantly created for humans across various domains. 
One significant challenge faced by many of these robots is their limited capability to navigate stairs, which hampers their functionality in multi-story environments. This study proposes a method aimed at addressing this limitation, employing reinforcement learning to develop a versatile controller applicable to a wide range of robots.
In contrast to the conventional velocity-based controllers, our approach builds upon a position-based formulation of the RL task, which we show to be vital for stair climbing. Furthermore, the methodology leverages an asymmetric actor-critic structure, enabling the utilization of privileged information from simulated environments during training while eliminating the reliance on exteroceptive sensors during real-world deployment. 
Another key feature of the proposed approach is the incorporation of a boolean observation within the controller, enabling the activation or deactivation of a stair-climbing mode. 
We present our results on different quadrupeds and bipedal robots in simulation and showcase how our method allows the balancing robot Ascento to climb 15cm stairs in the real world, a task that was previously impossible for this robot.

\end{abstract}

\section{INTRODUCTION}
Mobile ground robots have been widely studied and used for various tasks, such as delivery, inspection, and security,~\cite{rubio2019review}. 
While wheeled robots are efficient at traveling long distances, they lack the agility to traverse obstacles such as stairs. 
Legged systems, in contrast, are more agile, but lack efficiency and speed~\cite{biswal2021development}.
Hybrid wheeled-legged systems have been developed to harness the advantages of both technologies, but pose significant challenges for control due to their complex dynamics~\cite{bjelonic2022survey}.
Recently, \ac{RL} has emerged as a promising technique for controlling these complex robotic systems. In particular, interesting results have been achieved using \ac{RL} to develop controllers for wheeled-legged quadrupeds~\cite{lee2022control, vollenweider2022advanced}.

\begin{figure}
  \begin{minipage}[t]{0.24\textwidth}
  \centering
    \includegraphics[width = \textwidth,trim={16cm, 13cm, 23cm, 3cm}, clip]{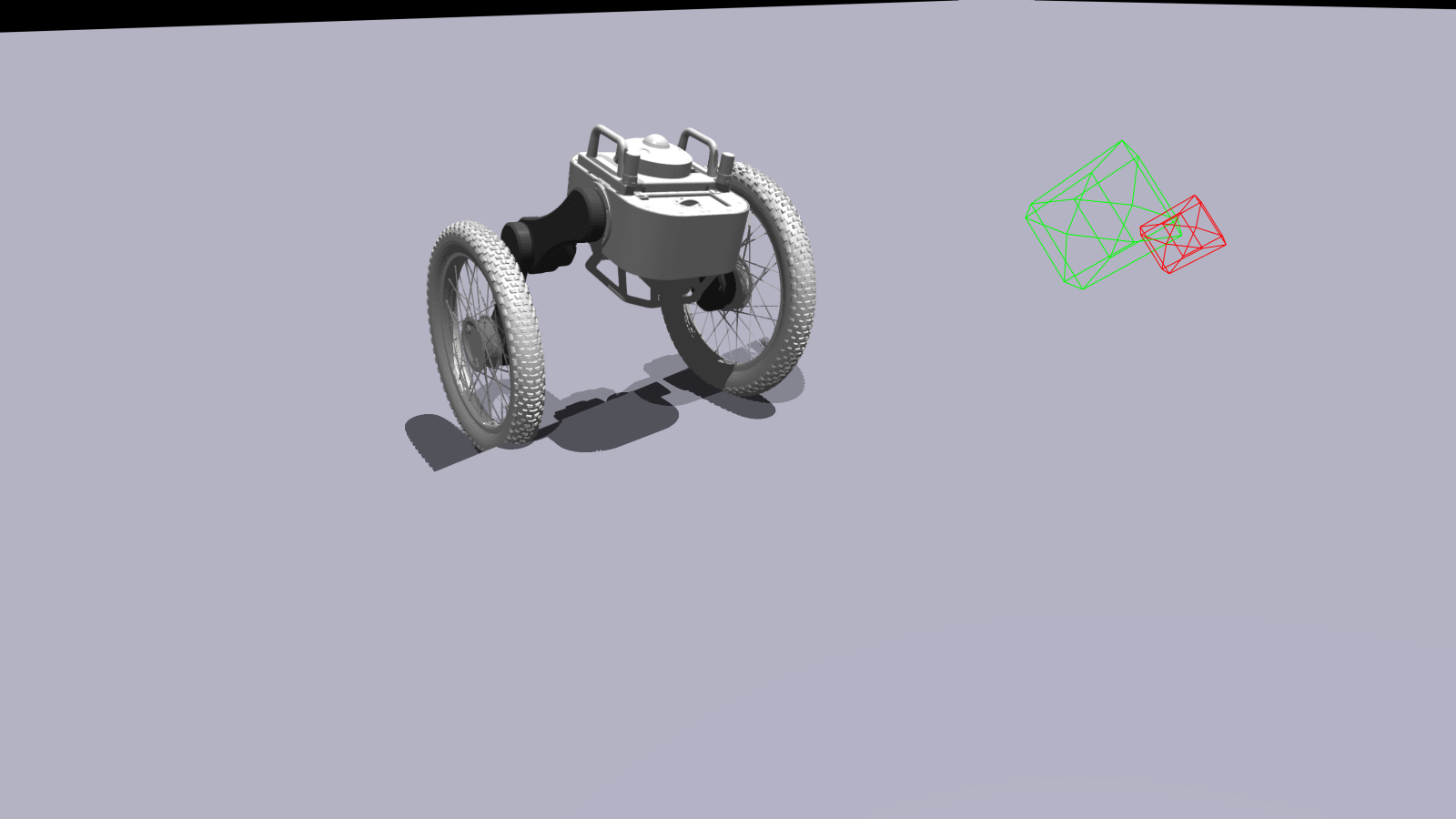}
    \vspace*{-0.3cm}
  \end{minipage}
  \hfill
  \begin{minipage}[t]{0.24\textwidth}
  \centering
    \includegraphics[width = \textwidth,trim={15cm, 13cm, 27cm, 2cm}, clip]{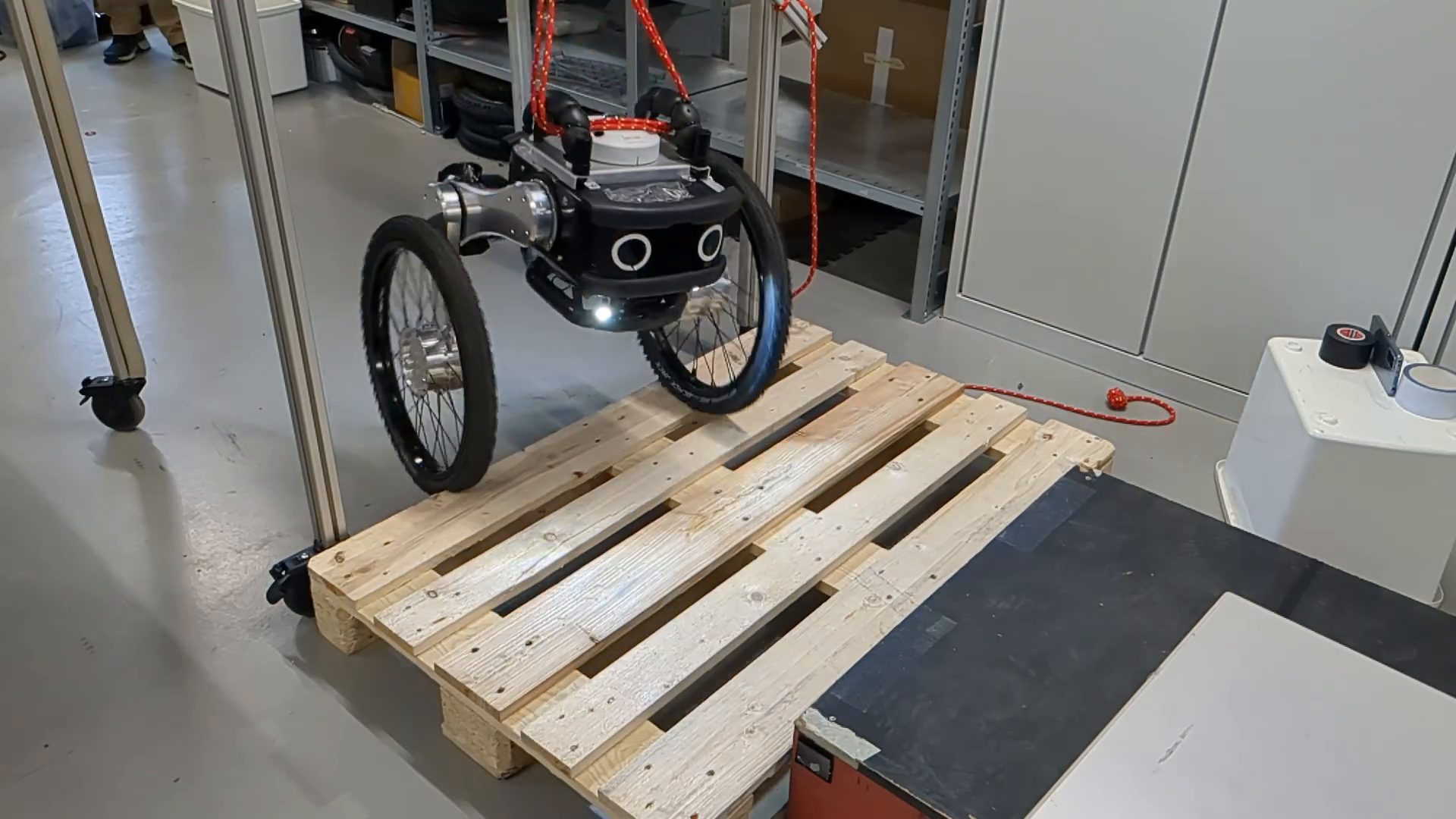}
    \vspace*{-0.3cm}
  \end{minipage}
  \hfill
  \begin{minipage}[t]{0.125\textwidth}
    \centering
    \includegraphics[width = \textwidth,trim={24cm, 11cm, 23cm, 11cm}, clip]{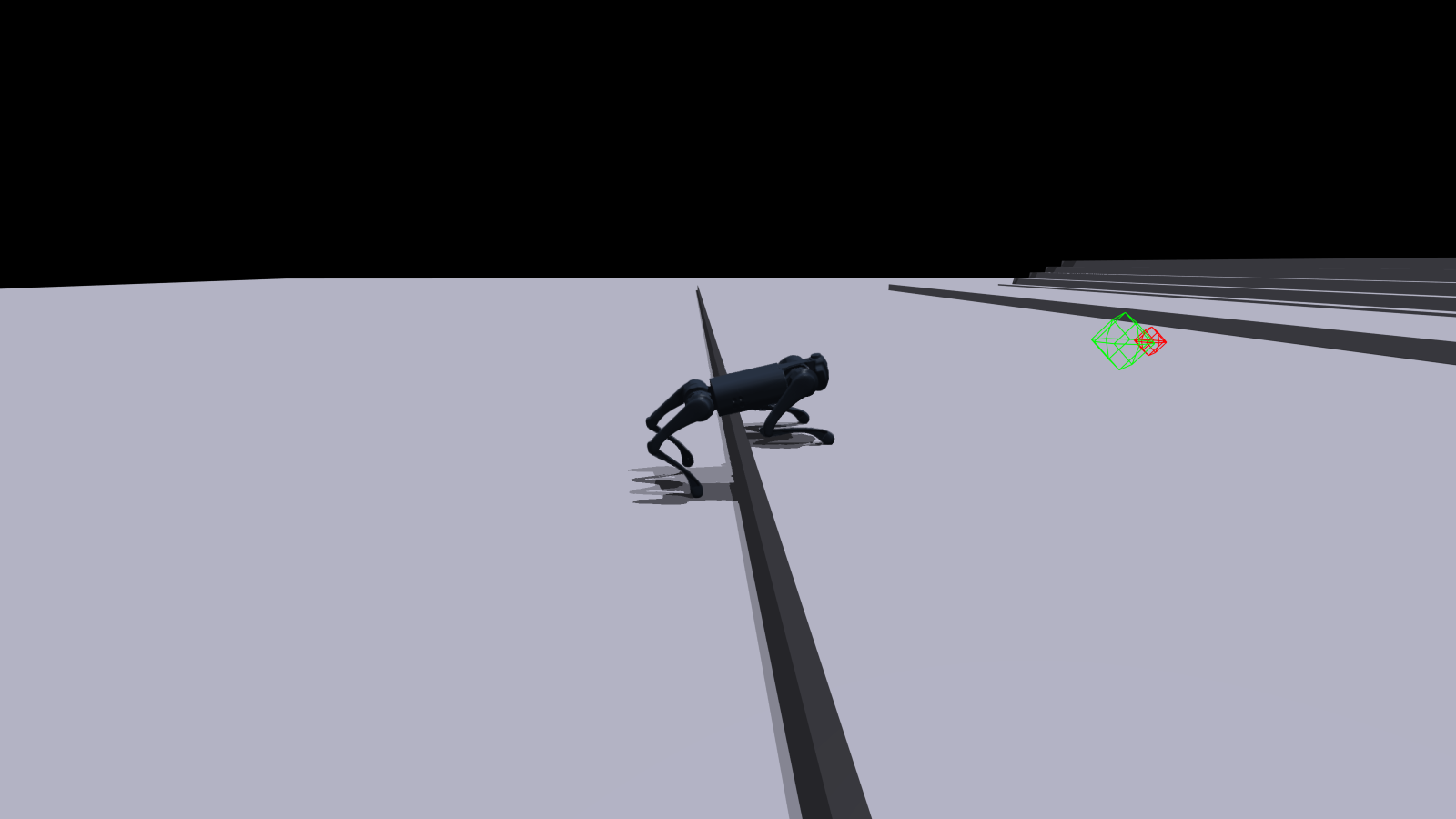}
  \end{minipage}
  \hfill
  \begin{minipage}[t]{0.095\textwidth}
    \includegraphics[width = \textwidth,trim={22cm, 10cm, 22cm, 5cm}, clip]{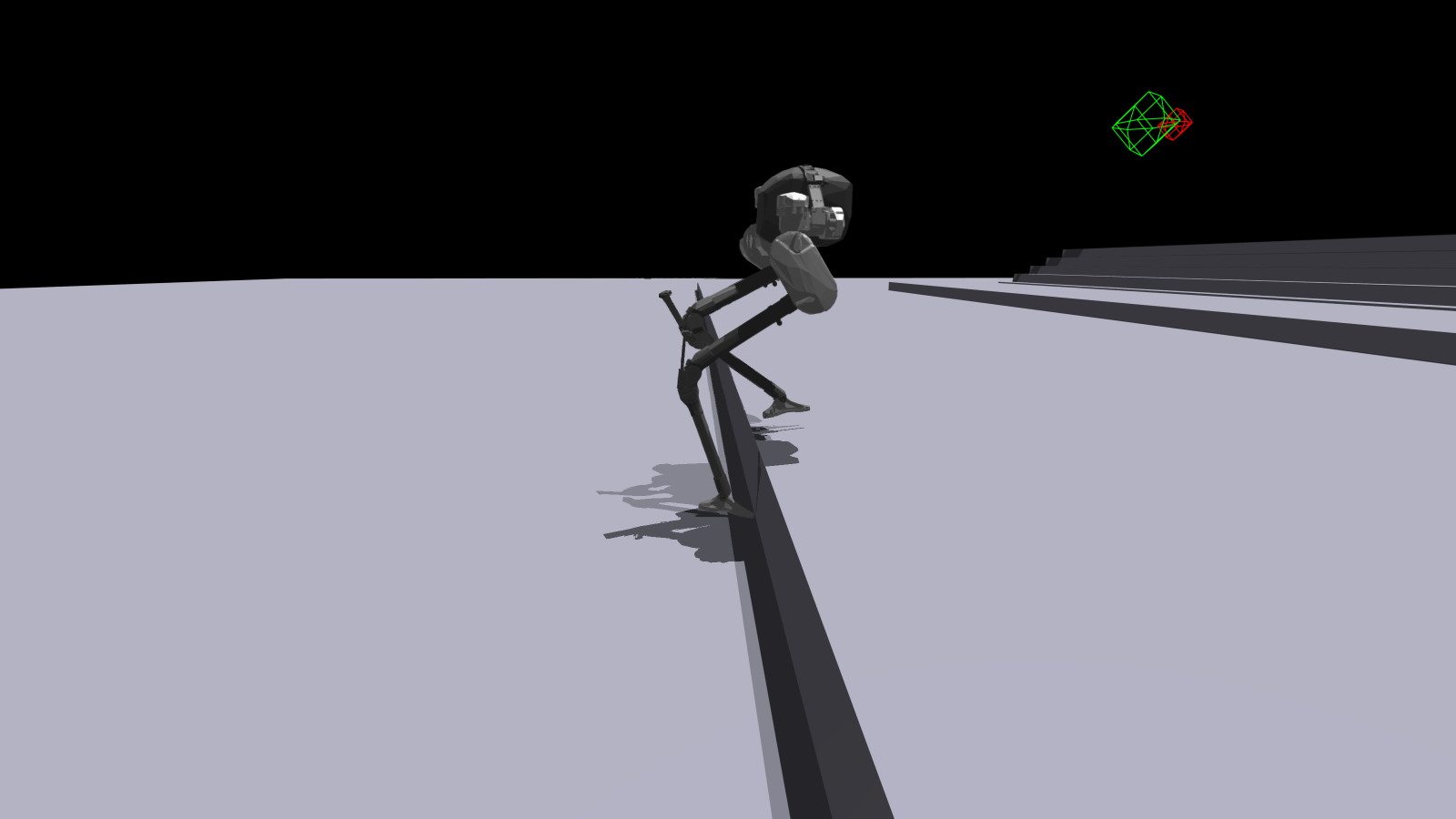}
  \end{minipage}
  \hfill
    \begin{minipage}[t]{0.09\textwidth}
    \includegraphics[width = \textwidth,trim={20cm, 7cm, 22cm, 4cm}, clip]{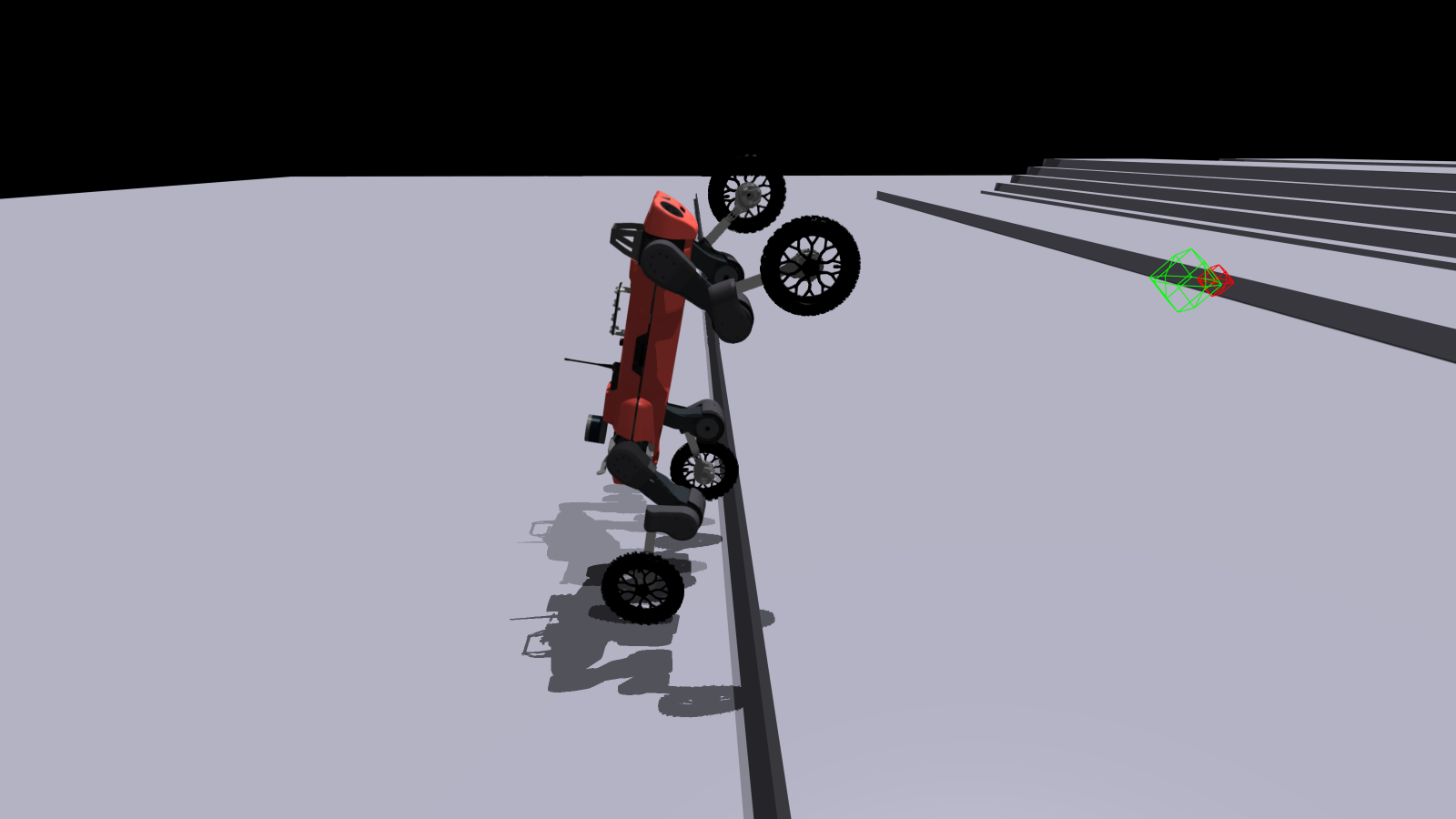}
  \end{minipage}
  \hfill
  \begin{minipage}[t]{0.155\textwidth}
    \includegraphics[width = \textwidth,trim={18cm, 11cm, 23cm, 8cm}, clip]{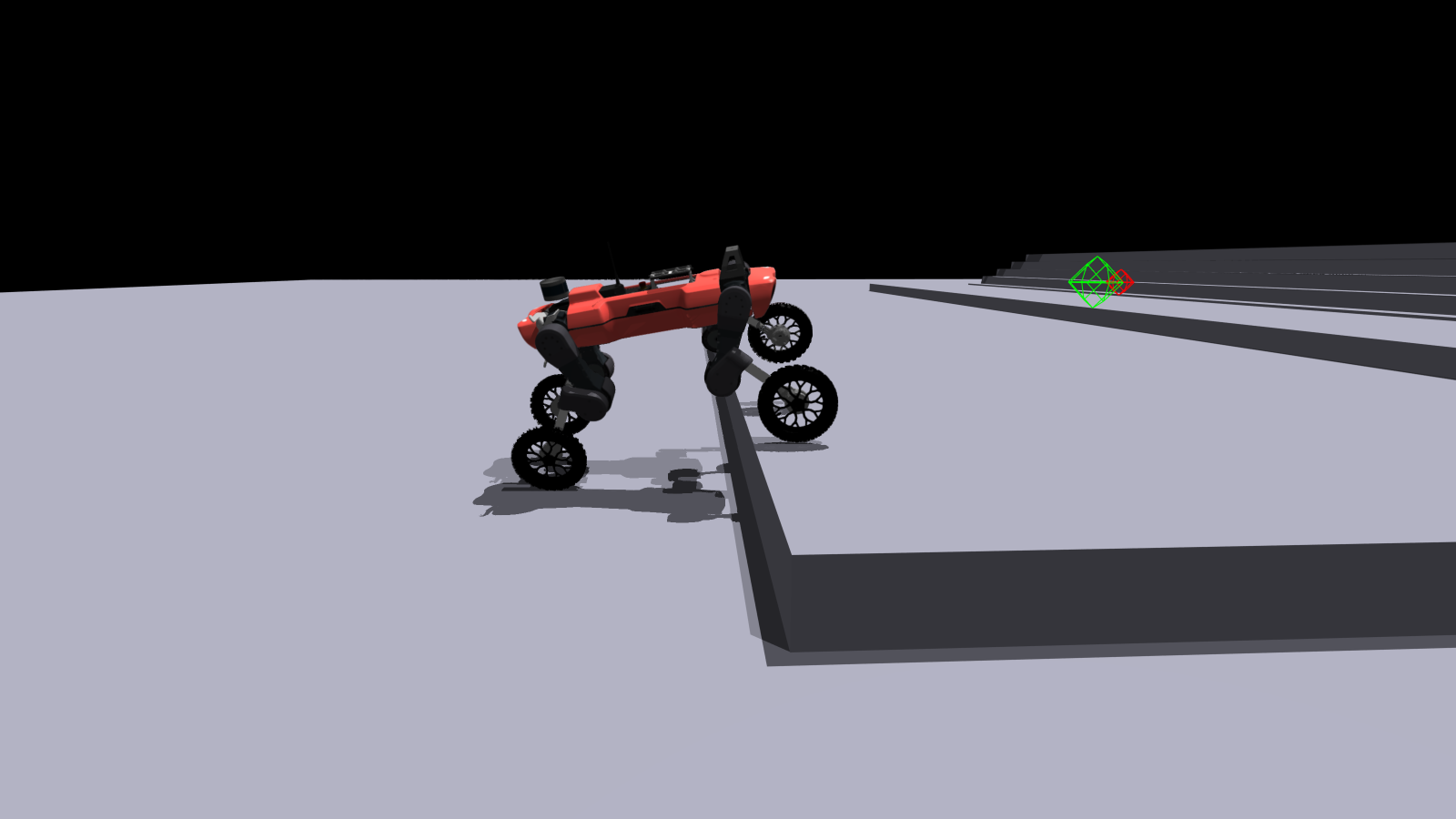}
  \end{minipage}
  \caption{\textbf{Proposed Method:} Ascento Robot, Unitree Go1, Cassie and ANYmal on Wheels climbing steps.}
  \label{fig:thumbnail}
\end{figure}

In this work, we study the problem of stair-climbing for legged and wheeled-legged robots. 
This is particularly difficult for bipeds since it requires a very dynamic movement where the robot keeps balance on only one point of contact while stepping up.
We propose a method to climb stairs with both legged and wheeled-legged robots, by developing step reflexes using privileged terrain information in an asymmetric actor-critic setup \cite{10160582}. 
Interesting reflex behaviors have been achieved on quadrupeds, presented by Lee et al. \cite{lee2020learning}.
Siekmann et al. have also shown that it is possible to blindly traverse stairs with the robot Cassie~\cite{siekmann2021blind}.
Additionally, we investigate the concept of a boolean mode switch for stair-climbing, thereby allowing a good performance on both regular terrain and stair ascent with the same control policy. 
Our controller operates with this boolean observation as its only exteroceptive information and no positioning system. 
Our main contributions are the following:
\begin{enumerate}
    \item We present an \ac{RL} task formulation to train policies capable of climbing stairs with quadruped robots, bipedal robots, and wheeled-legged balancing robots.
    \item Our method does not require any perceptive data or a positioning system such as \ac{SLAM} or \ac{GPS}, making it straight-forward to implement into a standard control stack.
    \item Our proposed system can successfully transfer to the real world and allows the wheeled bipedal robot Ascento to climb 15cm steps.
\end{enumerate}

\begin{figure*}[h]
   \centering
   \includegraphics[width=1.\textwidth,trim={9cm, 10cm, 8cm, 92cm}, clip]{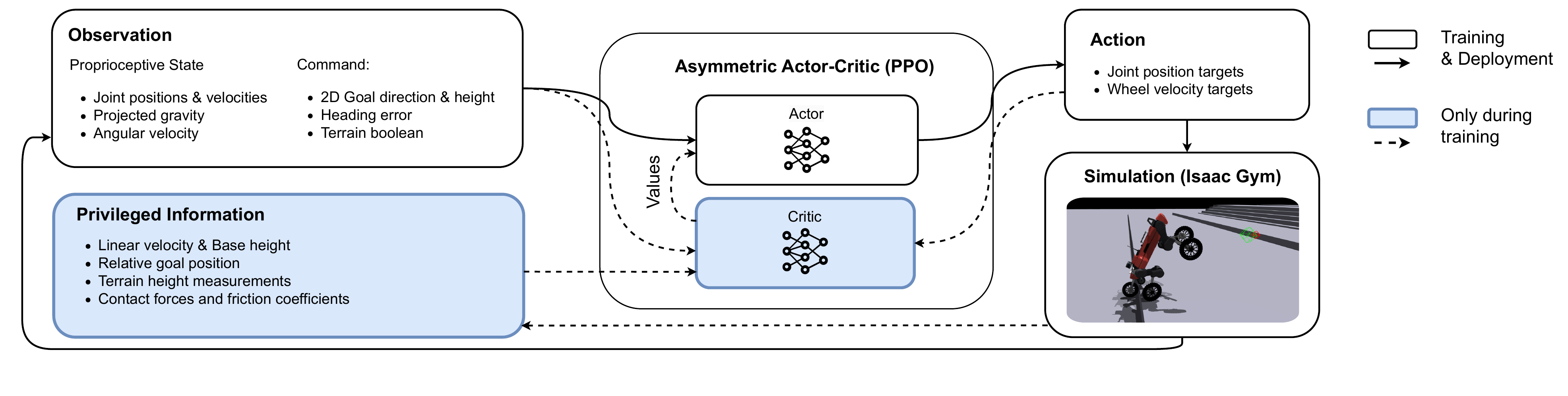}
   \caption{\textbf{System Overview during Training and Deployment:} At every training step, the algorithm receives the observation and privileged information. The actor outputs an action for the next simulation step. During deployment, the actor receives only the observation and outputs an action for the robot to execute.\vspace{-8pt}}
   \label{fig:overview}
\end{figure*}

\section{Literature Review}
\label{chap:lit-review}
\subsection{Legged Locomotion using \ac{RL}}
Model-based control methods that rely on optimization, such as \ac{MPC} and trajectory optimization, have seen widespread usage for legged locomotion problems in the past~\cite{wensing2022optimization}. However, these methods require a dynamics model of the system during execution leading to high complexity for high-dimensional systems.
Recently, \ac{RL} has emerged as a promising alternative for controlling legged systems showing unprecedented robustness.
Multiple recent works have successfully tackled locomotion for the ANYmal robot~\cite{hutter2016anymal} using model-free \ac{RL}.

In 2019, Hwuangbo et al.~\cite{hwangbo2019learning} proposed an \ac{RL} control policy using Trust Region Policy Optimization (TRPO)~\cite{schulman2015trust}, and an actuator network to model the motors more accurately in simulation. The policy is conditioned with the desired velocity of the base and outputs joint position targets. 
Another work, presented by Lee et al.~\cite{lee2020learning}, builds on this method and proposes a teacher-student architecture to include exteroceptive information during the training procedure. 
A teacher policy is trained using \ac{RL} and privileged information to output joint position targets. 
The student policy is then trained via imitation learning only with proprioceptive data as input.
Finally, in 2022, Miki et al.~\cite{miki2022learning} elaborated on the method to include proprioceptive and exteroceptive information via a belief encoder, allowing more efficient rough terrain navigation. 
Overall, the result of these works is a robust controller capable of navigating rough terrain reliably. 
However, one downside of these methods is the training time needed to learn a policy. 

Recently, Rudin et al.~\cite{rudin2022learning} proposed a new environment capable of simulating thousands of robots in parallel using a single Graphics Processing Unit (GPU). 
Using this framework and a standard Proximal Policy Optimization (PPO) implementation to learn a locomotion policy, they achieve stable results that transfer to the real world in under 20 minutes of training.
Additionally, Rudin et al.~\cite{rudin2022advanced} proposed an end-to-end policy for local navigation in challenging environments such as steep stairs and large gaps in terrain, overcoming the limitations of traditional path planning and locomotion control. 
The main contribution is the task formulation they propose, which gives the agent a position goal to reach by the end of the episode instead of a continuous velocity profile to track. 
This allows for a more flexible path and locomotion gait selection.
The approach enables robots to learn complex behaviors such as jumping.
We build upon this formulation and show it to be especially successful for obstacle crossing of robots that balance.
However, this method requires a positioning system and a mapping of the environment to obtain terrain information at deployment.
We address this limitation with our asymmetric actor-critic formulation.

\subsection{Hybrid Wheeled-Legged Locomotion}
Hybrid wheeled-legged robots aim to combine the agility benefits of legged robots with the efficiency of wheeled robots.
There has been a lot of work in the last few years to develop control algorithms for these robots using MPC and other methods such as trajectory optimization~\cite{bjelonic2019keep, bjelonic2020rolling, bjelonic2021whole,klemm2020lqr,klemm2024nonsmooth}.
However, these methods are very challenging to develop due to the complex nature of the highly nonlinear system dynamics at play. 
Similar to the case of purely legged robots, \ac{RL} has shown very promising results as an alternative method.
In 2022, Lee et al.~\cite{lee2022control} proposed a model-free \ac{RL} approach trained in simulation to learn a velocity-based controller. 
Furthermore, Vollenweider et al.~\cite{vollenweider2022advanced} proposed a framework to learn complex motions such as standing up on two wheels and sitting back down using motion priors and \ac{RL}. 
They also demonstrate the ability of the robot to balance on two wheels and follow velocity commands while maintaining this configuration.
However, this method is able to navigate only on flat terrain and very small obstacles.


\section{Methodology}
\label{chap:methodology}
Figure \ref{fig:overview} presents an overview of the proposed architecture during training and deployment. 
In the following sections, we present the learning environment used for training, the \ac{RL} task formulation, the training procedure, and our sim-to-real transfer efforts including deployment details.

\subsection{Learning Environment}
\label{sec:env}
\subsubsection{Simulator}
We used Isaac Gym~\cite{makoviychuk2021isaac} as our training platform, specifically designed for \ac{RL} applications.
Compared to other widely-used simulators such as Raisim~\cite{hwangbo2018per} and MuJoCo~\cite{todorov2012mujoco}, Isaac Gym's GPU-accelerated architecture enables significantly faster training of agents, thanks to its high parallelization. 
Isaac Gym supports domain randomization techniques~\cite{tobin2017domain}, which can enhance the robustness of \ac{RL} agents by introducing variations in the environment during training, thus aiding in the transfer of learned policies to the real world.

\begin{figure}[t]
  \begin{minipage}[t]{0.15\textwidth}
  \centering
    \includegraphics[width = \textwidth]{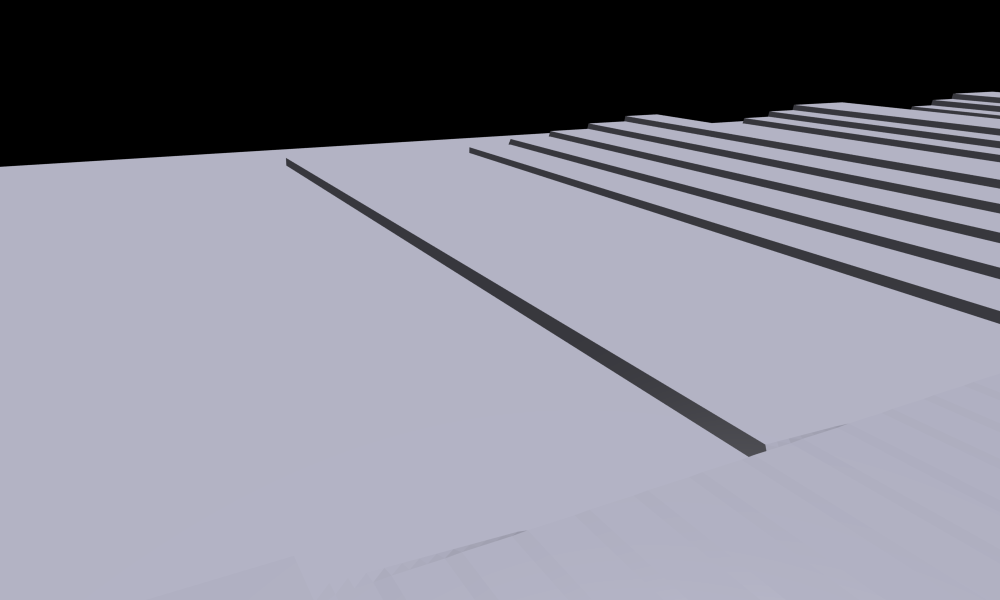}
    \vspace*{-0.2cm}
  \end{minipage}
  \hfill
  \begin{minipage}[t]{0.15\textwidth}
    \centering
    \includegraphics[width = \textwidth]{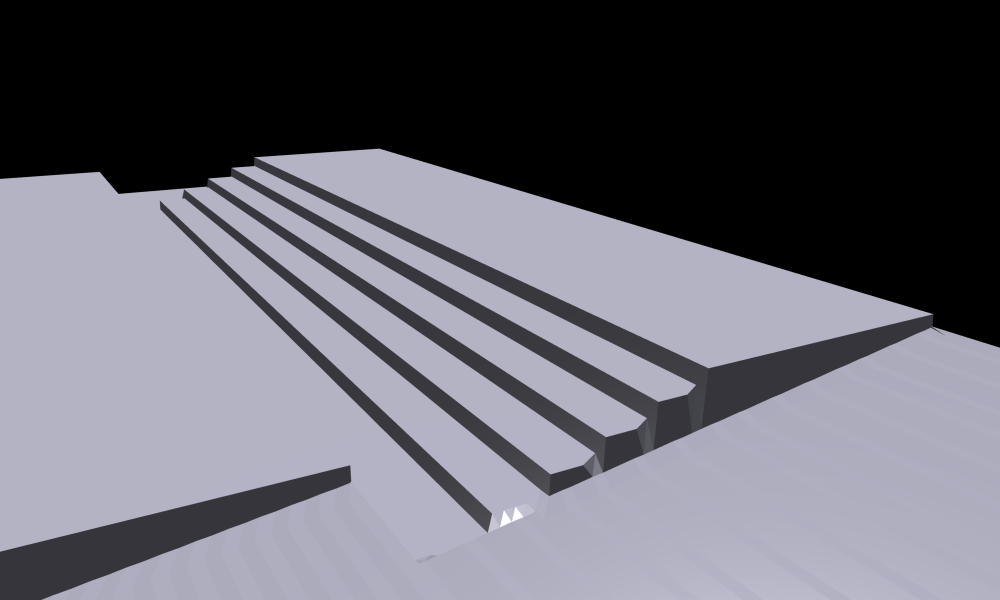}
    \vspace*{-0.2cm}
  \end{minipage}
  \hfill
  \begin{minipage}[t]{0.15\textwidth}
    \includegraphics[width = \textwidth]{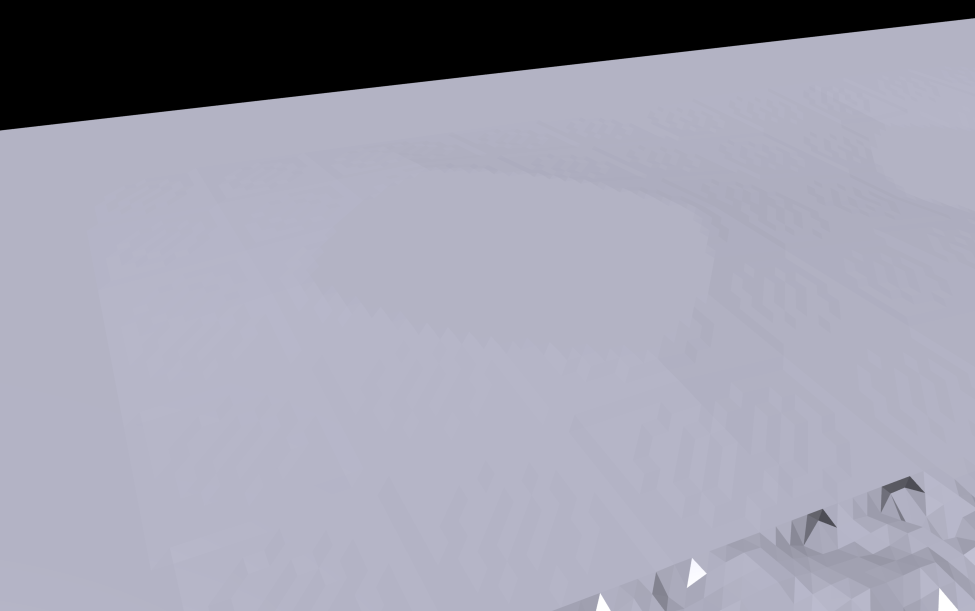}
    \vspace*{-0.2cm}
  \end{minipage}
  \hfill
  \begin{minipage}[t]{0.15\textwidth}
    \includegraphics[width = \textwidth]{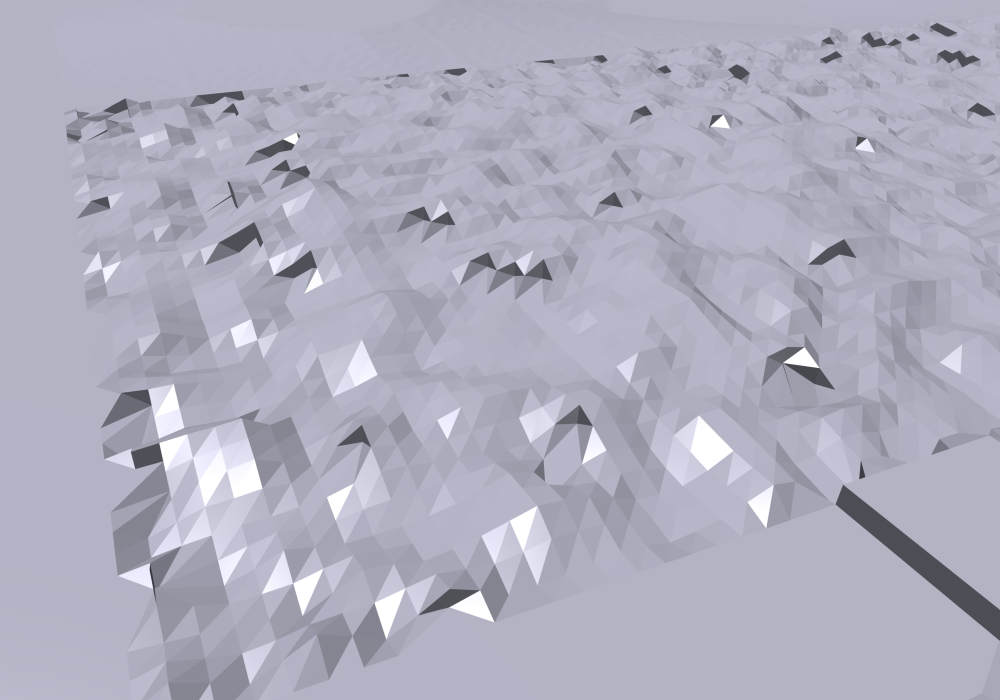}
  \end{minipage}
  \hfill
  \begin{minipage}[t]{0.15\textwidth}
    \includegraphics[width = \textwidth]{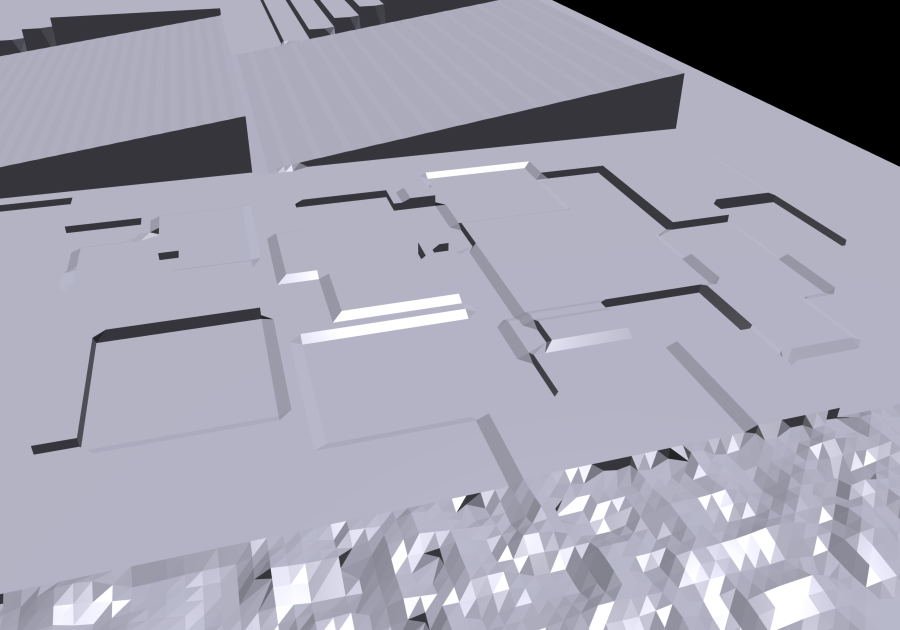}
  \end{minipage}
  \hfill
  \begin{minipage}[t]{0.15\textwidth}
    \includegraphics[width = \textwidth]{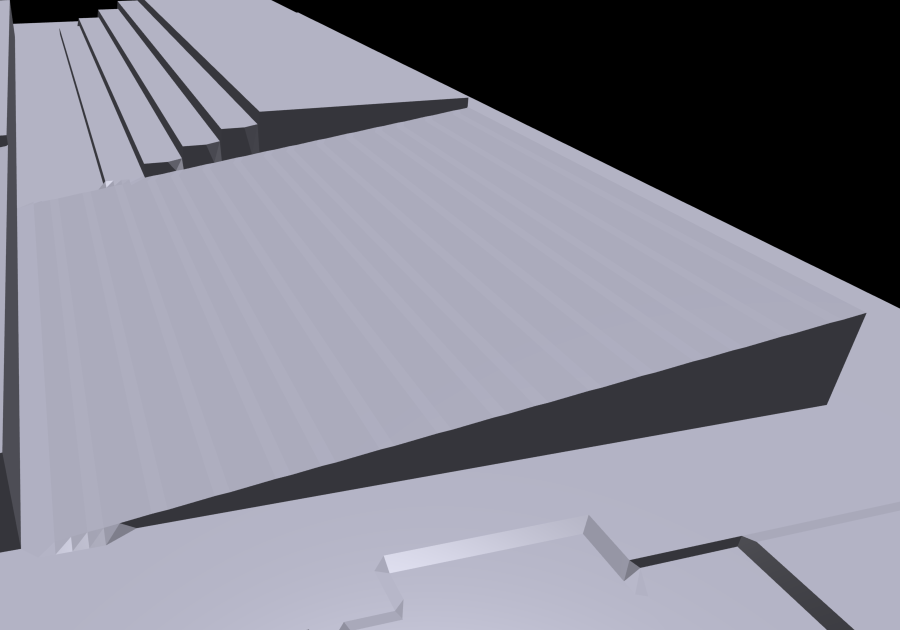}
  \end{minipage}
  \caption{\textbf{Different Training Terrains:} First row left to right: Single step, staircase, smooth pyramid. Second row: rough pyramid, discrete obstacles, smooth slope.}
  \label{fig:terrains}
\end{figure}

\subsubsection{Learning Algorithm}
We use the \ac{RL} algorithm PPO, with an asymmetric actor-critic structure~\cite{schulman2017proximal}. 
This is a variant of the standard actor-critic algorithm in \ac{RL}, where the actor and critic use separate networks, allowing them to update independently~\cite{mnih2016asynchronous, 6392457}. 
In this work, we use the open-source PPO implementation published by the Robotics Systems Lab (RSL) from ETH~\cite{rudin2022learning}\footnote{https://github.com/leggedrobotics/rsl\_rl}.
We remove the bootstrapping element from this implementation since we deal with a finite horizon task formulation \cite{rudin2022advanced}. 
All of our experiments converge after \(~\)5k learning iterations.
For all of our training and experiments, we used a single GPU, an Nvidia DGX A100 with 16 GB of memory. 

\subsubsection{Terrains}
\label{sec:terrains}
Our environment is divided into 6m $\times$ 6m terrains.
There are six columns: two are stairs and there is one of each of the other terrains. 
There are twelve rows for increasing curriculum difficulty. 
We have noted that the policies are more robust when exposed to a bigger variety of obstacles. 
We use the following types of terrain, shown in figure \ref{fig:terrains}:

\textbf{Stairs: }The stairs columns are divided into two sections. The first half, i.e. six rows, are terrains with one step of gradually increasing height. There is space before and after the step. These terrains are meant for the agent to learn the basic motion of going up a step before tackling continuous stairs terrains. The second half of the rows contain continuous stair flights of five steps. Step heights and widths vary depending on the robot we are training.

\textbf{Smooth slopes: } Smooth slopes are slanted planes. These are especially important when training wheeled robots to reduce drift, forcing them to hold a stable position. The slopes range from 0 degrees to 8.5 degrees.

\textbf{Discrete obstacles: }The discrete obstacles are random rectangular obstacles created on flat terrain. Their height goes up to 10cm and their dimensions in length and width vary between 1m and 2m.

\textbf{Smooth \& rough pyramid slopes: }These environments are pyramids with a flat platform in the middle. Their inclination goes from 0 to 8.5 degrees. In the case of the rough pyramids, they have randomized noise in the terrain height of up to 5cm, which gives them a bumpy surface.

\subsubsection{Robot Model}
We evaluate our method on four different robots: The Unitree Go1 quadruped~\cite{go1}, the bipedal robot Cassie~\cite{Cassie}, the wheeled-legged balancing robot Ascento~\cite{klemm2019ascento} and the wheeled-legged quadruped ANYmal on Wheels~\cite{bjelonic2019keep}, illustrated in figure \ref{fig:thumbnail}. 
For ANYmal on Wheels, we conduct experiments in quadruped mode and biped mode.

\begin{figure*}[h]
 \centering
 \begin{minipage}[t]{0.34\textwidth}
  \centering
    \includegraphics[width = 0.94\textwidth, trim={3.3cm, 0cm, 1.3cm, 0cm}, clip]{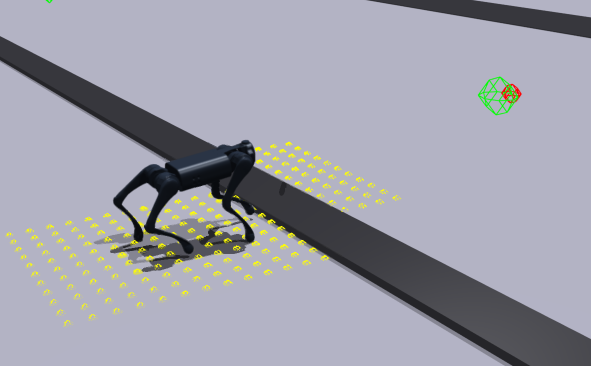}
  \end{minipage}
  \hfill
  \begin{minipage}[t]{0.65\textwidth}
    \centering
    \includegraphics[width = 0.93\textwidth, trim={2cm, 4cm, 2cm, 5.5cm}, clip]{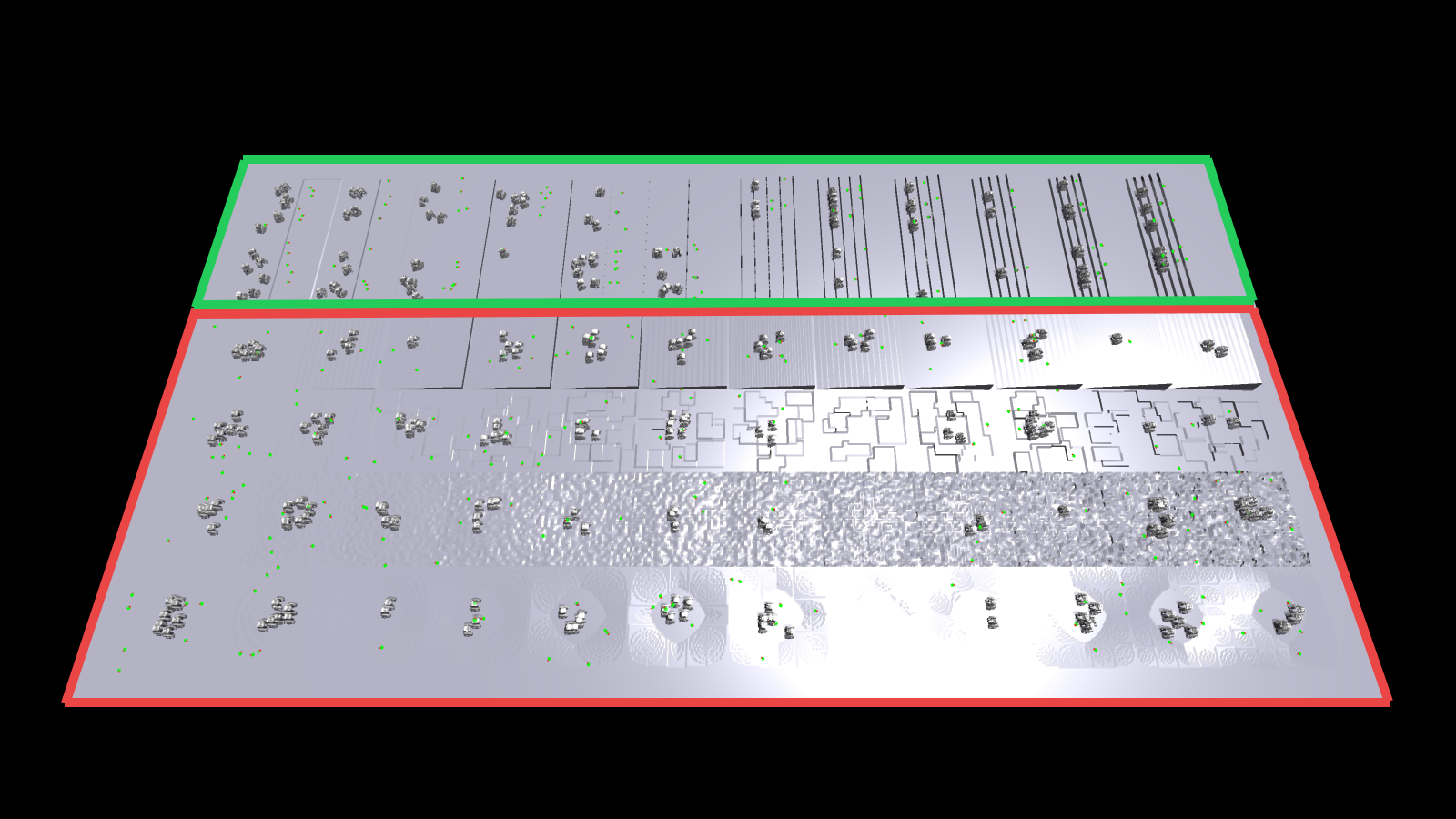}
  \end{minipage}
  \caption{\textbf{Task setup:} a) The goal pose is marked in green and red. The yellow dots represent the terrain height measurements which are part of the privileged information about the terrain. b) Training curriculum: Different training terrains, see section \ref{sec:terrains}. Progressively harder from left to right. The stairs terrains, where the terrain boolean is set to 1 during training, are delimited in green. In the other terrains, this observation is set to 0.\vspace{-8pt}}
  \label{fig:task}
\end{figure*}

\subsection{Task Formulation}
We formulate our \ac{RL} task as a position-based navigation task, inspired by Rudin et al.'s~\cite{rudin2022advanced}. 
The episodes last 6 seconds ($T = 6s$), and to solve the task, the robot's base needs to be at a goal location before the end of the episode.
For the step and stairs terrains, the robot always spawns at the bottom of the steps, 0.3 to 1m away from them and roughly facing them ($\pm0.1rad$). 
The goal is sampled at the top and also has a heading within the same range.
For the other terrains, the robot spawns in the middle of the terrain and a goal is sampled within a 1m radius with a random orientation.
Figure \ref{fig:task} illustrates our \ac{RL} task in a step environment.

\subsubsection{State}

\begin{table}
\setlength{\tabcolsep}{3pt}
\begin{center}
 \caption{Observation \& Privileged Information.}\vspace{1ex}
 \label{tab:obs}
 \begin{tabular}{ll|cccc}
 \hline
 Symbol                 & Observation           & Units             & Coeff.            & Size  & Noise (\%)\\ 
 \hline \hline
 $\vec{\dot{\theta}}$   & Angular velocity      & rad/s             & 0.25              & 3     & $\pm 20$\\
 $\vec{\gamma}$         & Projected gravity     & N/A               & 1.0               & 3     & $\pm 5$\\
 $\vec{g_{dir}}$        & Direction to goal     & N/A               & 1.0               & 2     & $\pm 0$\\
 $\theta_{err}$         & Heading error         & rad               & 1.0               & 1     & $\pm 0$\\
 $h_{target}$           & Height command        & m                 & 1.0               & 1     & $\pm 0$\\
 $b$                    & Terrain boolean       & N/A               & 1.0               & 1     & $\pm 0$\\
 $\vec{q}$              & Joint positions       & rad               & 1.0               & 4     & $\pm 1$\\
 $\vec{\dot{q}}$        & Joint velocities      & rad/s             & 0.05              & 6     & $\pm 150$\\
 $\vec{a_{last}}$       & Last action           & rad \& rad/s      & 1.0               & 6     & $\pm 0$\vspace{0.1cm}\\
 \hline
  Symbol                 & Privileged Information           & Units             & Coeff.            & Size  & Noise (\%)\\
\hline
\hline
  $v_x$                 & Linear velocity           & m/s   & 0.25              & 1             & - \\
 $h$                    & Base height               & m     & 1.0               & 3             & -\\
 $\vec{g_{rel}}$        & Relative goal position    & m     & 1.0               & 3             & -\\
 ${\mathbf{H_{terrain}}}$  & Terrain height         & m     & 5.0               & 187           & -\\
 $\vec{f_{wheels}}$     & Contact forces            & N     & 0.01              & 6             & -\\
 $\mu$                  & Friction coefficient      & N/A   & 1.0               & 1             & -\\
 \hline
 \end{tabular}
\end{center}
\end{table}

We use an asymmetric actor-critic structure, so we divide our state into two: an observation, to which both the actor and critic have access, and privileged information, which only the critic can access during training.
Table \ref{tab:obs} presents the observations used by the actor during training and deployment as well as the privileged information.
First, privileged information contains the linear velocity and absolute base height of the robot. 
These values are not straightforward to obtain on the real robots, which is why they are not included in the actor's observation.
It also contains the goal relative position in $x, y, z$ instead of only a 2D direction in the $x, y$-plane.
Finally, we provide some ground truth information from the simulator, such as terrain height measurements, contact forces, and friction coefficient of the terrain.
This allows the critic to locate the stairs and provide better guidance to the actor.
The height measurements are sampled from a $1.6 \times 1.0$m grid around the robot and spaced equally. 
The contact forces are filtered using an averaging sliding window with a window of 5 timesteps, equivalent to 0.1 seconds.
Additionally, the observation contains uniform noise for the actor whereas it does not for the critic.

\subsubsection{Actions}
The actions are a vector of robot-dependent dimensions.
For the articulated joints, the commands are directly interpreted as goal positions to be tracked by a low-level PD controller. Analogously,  actions are interpreted as target angular velocities for the wheeled joints.

\subsubsection{Rewards}
\label{sec:rewards}
Our task-specific rewards are presented in table \ref{tab:rewards}. 
The main idea behind the position-based task formulation is to give more freedom to the actor to explore different strategies to climb steps and find a good solution that might have been difficult to obtain by manual crafting. 
Therefore, the main reward of our task, reward \#1, only comes into play during the last 2 seconds of the episode ($T_r = 2s$).
This reward encourages the robot to be at the goal location at the end of the episode without adding any further constraints.
To aid convergence, reward \#2 is a temporary reward that is given to the agent until it reaches half of the best possible performance for reward \#1. 
It gives an incentive for the robot to have a movement toward the goal.
Reward \#3, on the other hand, penalizes stalling unless the robot is already close to the goal.
Finally, reward \#4 encourages the robot to face the goal.
In addition to these rewards, we use standard robot-specific shaping rewards to encourage smooth and stable motions such as rewarding feet air time and penalizing joint acceleration for legged robots~\cite{rudin2022learning}. 

\begin{table}[h]
\begin{center}
\setlength{\tabcolsep}{4pt}
 \caption{Rewards.}\vspace{1ex}
 \label{tab:rewards}
 \begin{tabular}{l|ccc}
 \hline
 \# & Reward                  & Formula                                                                                             & Coefficient\\ \hline \hline
 1  & $r_{position}$          & $\frac{1}{T_r} \frac{1}{1 + \|\vec{x} - \vec{r_{goal}}\|^{2}}$ if $t > T - T_r$                         & 10.0 \\
 2  & $r_{pos\_bias}$         & $\frac{\dot{\vec{x}} \cdot (\vec{r_{goal}} - \vec{x})}{\|\dot{\vec{x}}\| \|\vec{r_{goal}} - \vec{x}\| }$    & 1.0 \\
 3  & $r_{stall}$             & \begin{tabular}{@{}c@{}}$-1$ if $\|\dot{\vec{x}}\| < 0.1m/s$ \\ and $\|\vec{x} - \vec{r_{goal}}\| > 0.5m$\end{tabular} & 1.0 \\
 4  & $r_{face\_goal}$        & $-\| \theta - \theta_{goal} \|$ if $\|\vec{x} - \vec{r_{goal}}\| > 0.5m$                                & 0.1 \\ 
 \hline
 \end{tabular}
\end{center}
\end{table}


\subsubsection{Curriculum}
We spawn 4096 robots in the environment, equally distributed among the terrain types and starting in the easiest levels.
We use the game-inspired curriculum from \cite{rudin2022learning}.
If the robot is within 20cm of the goal, we move it one level up, and if the robot is more than 50cm away from the goal, we move it down, unless it is already at the lowest level.
If a robot reaches the highest level, we proceed to a random assignment to avoid catastrophic forgetting.
Figure \ref{fig:task} illustrates a top view of the environment.

\subsection{Sim-to-real Transfer}\label{sec:sim-to-real}
During training, we use domain randomization to facilitate transfer to the real world.
First, we randomize friction with the terrain. 
For wheeled-legged robots, we noticed that if friction is too high, the robot learns to use its grip between the wheels and the step to go up. 
While this strategy works with the contact model of the simulator, it does not transfer due to unmodelled tire dynamics. In the real world, kick-back and irregularities in the tire profile can cause slippage, as the wheel momentarily loses contact with the step, further discussed in \ref{sec:real-world-results}
Furthermore, we used standard domain randomization techniques from literature \cite{8202133, tan2018sim} and applied random ``pushes" to the robot by assigning a random base linear velocity sporadically throughout training.
The velocities vary between 0 and 0.5 m/s and are applied at most every 3 seconds.
This helps with overall robustness to disturbances and deviations from desired trajectories.
Finally, it was particularly important to simulate delays in the control loop~\cite{8593894}.
We achieve this by randomly using states from the previous time step with a probability of 50\% to compute the subsequent actions. 
This simulates a \SI{20}{ms} delay.

\subsubsection{Real-World Deployment}
In contrast to deployment in simulation, where the observations can directly be obtained from the simulator, we need to measure and estimate the required quantities for real-world deployment. We deploy our policy on the newest Ascento robot prototype.
During real-world deployment, we no longer need privileged information nor the critic. 
To control the robot with the learned policy, i.e. the actor network, we reconstruct the observation as follows:

\textbf{Proprioceptive state: }The proprioceptive state includes joint information as well as orientation information. 
    The joint positions are measured from joint encoders, and the joint velocities are calculated numerically from joint position change during the last control iteration.
    Angular velocities and projected gravity are calculated from Inertial Measurement Unit (IMU) data.

\textbf{Command: }The command includes the goal direction, base height target, heading error, terrain boolean, and last actions.
    For our experiments, we use a remote controller to provide all these commands, except the last actions, which are stored in memory.
    Commanding the forward position and heading errors behaves similarly to controlling a velocity-based controller in terms of user experience, with the added benefit that the robot can slightly deviate from the commands to successfully overcome obstacles.


\section{Experiments \& Results}
\label{chap:results}
\subsection{Simulation Results}
\begin{table}
        \centering
        \caption{Simulation Results for Step Climbing.}
        \label{tab:results}
        \setlength{\tabcolsep}{3pt}
         \begin{tabular}{l|l|c|c|c|c|c|c}
         \hline
         Robot                          & Exp.                          & \multicolumn{6}{c}{Success Rate (\%)}\\
        \hline \hline
        Step height                    &                                & 14cm      & 18cm      & 22cm      & 26cm      & 30cm      & \begin{tabular}{c}Other\\terrains\end{tabular}\\
        \hline     
        
         \multirow{4}{*}{\begin{tabular}{l}Unitree\\
         Go1\end{tabular}}              & Bool on                       & 99.7      & 99.7      & 98.5      & 43.1      & *0.0      & 97.7 \\
                                        & Bool off                      & 98.8      & 98.3      & 93.5      & 19.1      & *0.0      & 98.6  \\
                                        & No bool                       & 100.0     & 99.8      & 99.6      & 56.4      & *0.0      & 98.7  \\
                                        & No priv.                      & 15.4      & 12.9      & 8.8       & 3.9       & *1.0      & 95.3  \\  
        \hline
         \multirow{4}{*}{\begin{tabular}{l}Cassie
         \end{tabular}}                 & Bool on                       & 100.0     & 100.0     & 99.7      & 99.0      & 98.3      & 98.8  \\
                                        & Bool off                      & 97.4      & 92.2      & 87.8      & 78.2      & 64.7      & 98.1  \\
                                        & No bool                       & 99.2      & 97.2      & 90.6      & 79.6      & 74.3      & 69.4  \\
                                        & No priv.                      & 0.0       & 0.0       & 0.0       & 0.0       & 0.0       & 22.3  \\
        \hline
         \multirow{4}{*}{\begin{tabular}{l}Ascento\
         \end{tabular}}                 & Bool on                       & 87.6      & 82.3      & 69.4      & *31.4     & *1.4      & 97.7 \\
                                        & Bool off                      & 23.2      & 18.6      & 9.6       & *0.1      & *0.0      & 98.1 \\
                                        & No bool                       & 87.9      & 82.5      & 47.9      & *4.4      & *0.0      & 62.5 \\
                                        & No priv.                      & 0.0       & 0.0       & 0.0       & *0.0      & *0.0      & 80.2\\
        \hline
         \multirow{4}{*}{\begin{tabular}{l}ANYmal\\
         on Wheels\\
         (Quadruped)\end{tabular}}      & Bool on                       & 93.2      & 93.8      & 90.2      & 61.1      & 18.7      & 95.9\\
                                        & Bool off                      & 73.8      & 40.8      & 18.9      & 4.8       & 0.5       & 97.7\\
                                        & No bool                       & 34.3      & 37.9      & 26.4      & 34.4      & 14.8      & 20.9\\
                                        & No priv.                      & 0.3       & 0.0       & 0.0       & 0.0       & 0.0       & 62.7\\
        \hline
         \multirow{4}{*}{\begin{tabular}{l}ANYmal\\
         on Wheels\\
         (Biped)\end{tabular}}          & Bool on                       & 93.9      & 92.8      & 93.6      & 93.6       & 93.4      & 85.0 \\
                                        & Bool off                      & 71.0      & 45.4      & 27.8      & 14.2       & 6.2       & 94.9\\
                                        & No bool                       & 68.9      & 70.0      & 58.2      & 35.0       & 34.4      & 85.7 \\
                                        & No priv.                      & 0.0       & 0.0       & 0.0       & 0.0        & 0.0       & 78.8 \\
        \hline
        \multicolumn{8}{l}{\textbf{Experiments: }\textit{Bool on: }Our proposed method with terrain bool set to 1. }\\ 
        \multicolumn{8}{l}{\textit{Bool off: }Our proposed method with terrain bool set to 0.}\\ 
        \multicolumn{8}{l}{\textit{No bool: }Ablation trained without a terrain boolean observation.}\\
        \multicolumn{8}{l}{\textit{No priv.: }Ablation trained without privileged information.}\\
        \multicolumn{8}{l}{\textbf{Legend: }\textit{Success} means the robot reached the goal $\pm$20cm.}\\
        \multicolumn{8}{l}{\textbf{*Note: }This height exceeds the step height the robot was trained on.}
         \end{tabular}
\end{table}

We evaluate our method on all robots and conduct an ablation study to showcase the importance of the asymmetric actor-critic structure using privileged information as well as the terrain boolean observation. 
Table \ref{tab:results} shows the success rate obtained for each robot for climbing a single step. 
For each experiment, 2000 trials are conducted per robot.
First, we conclude that the privileged information is critical for successfully learning the task.
We attribute this mainly to the terrain information, which helps guide the actor to a terrain-cautious behavior.
On the other hand, the proposed boolean observation is particularly important for bipedal robots because it allows them to learn distinct skills for stair-climbing while learning a robust general policy to navigate the other terrains. 
This is further discussed in section \ref{sec:bool}.

\subsubsection{Position Control Analysis}
\begin{figure}
\centering
\begin{minipage}[t]{0.45\textwidth}
      \centering
        \includegraphics[width = \textwidth, trim={0, 8cm, 10cm, 7cm}, clip]{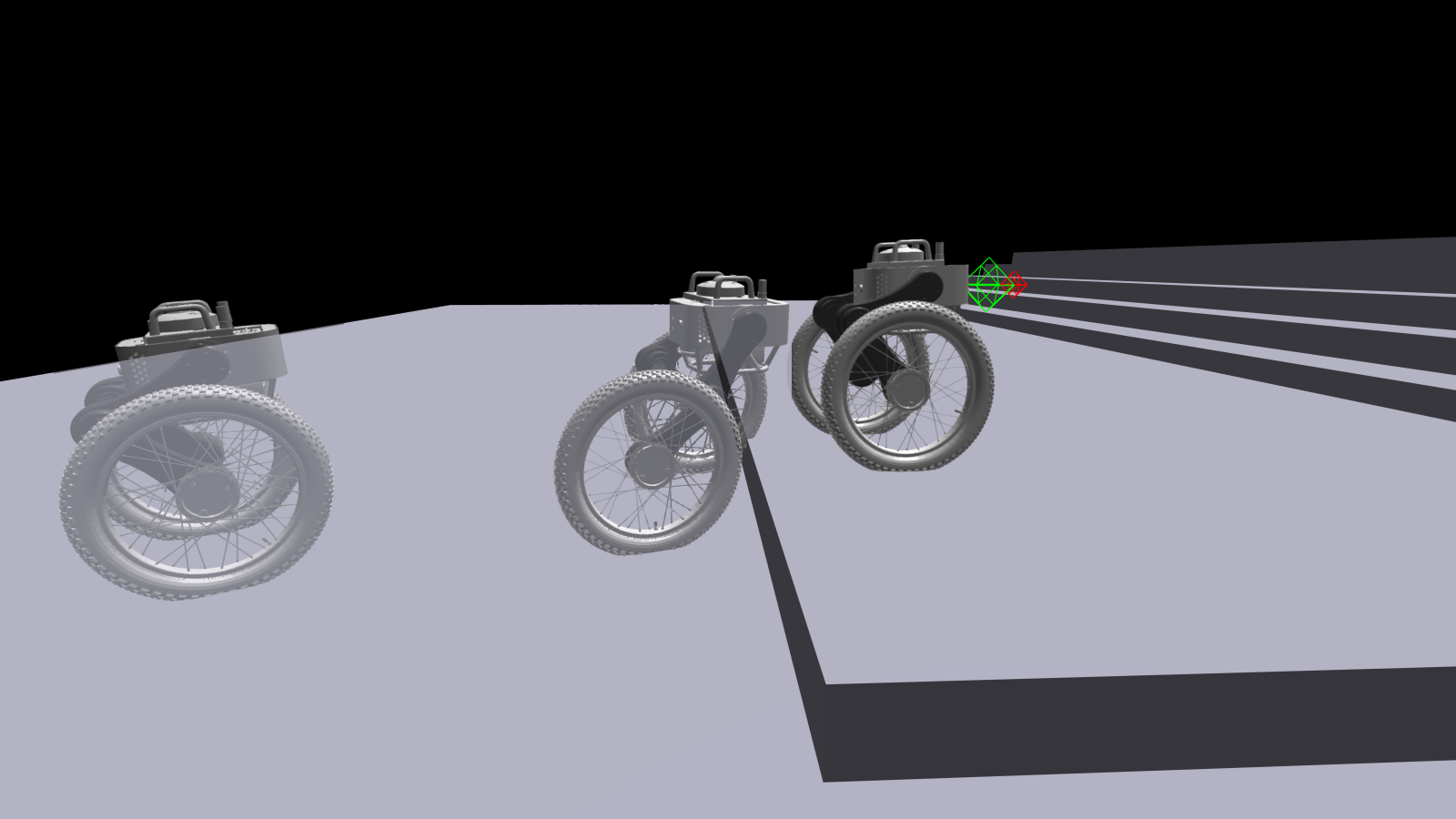}
      \end{minipage}
      \hfill
      \begin{minipage}[t]{0.45\textwidth}
        \centering
        \includegraphics[width = \textwidth]{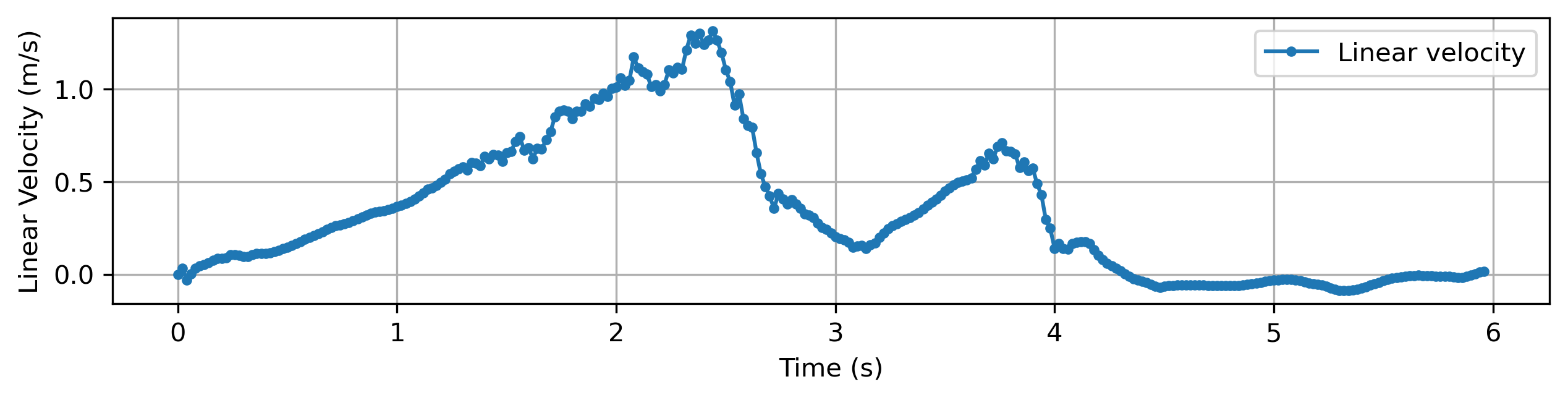}
    \end{minipage}
\caption{\textbf{Step motion:} Velocity profile of the step-up motion with our method.}
\label{fig:vel-position}
\end{figure}
All the trained robots achieve a success rate higher than $75\%$ on standard height steps of 18cm with our method.
Figure \ref{fig:vel-position} shows the velocity profile of the Ascento robot using our policy when going up a 20cm step in simulation as an example.
We can see that the controller adopts a very non-linear velocity profile. 
The robot accelerates at the beginning of its trajectory to approach the goal. 
However, when it reaches the step, it decelerates and performs the learned step-up motion, before accelerating again to reach the goal.
We were not able to obtain this kind of behavior with pure velocity-based approaches where the robot is constrained to follow a specific velocity.
Here, we only provide a direction toward the goal and the robot is free to adjust its velocity and path to overcome the step that is in the way.
Additionally, it is important to keep in mind that the controller does not use any perception other than the boolean observation, and it has no knowledge of the exact location of the goal.
The detection of the step happens through proprioception. 
The robot Ascento learns to advance with one leg slightly in front of the other, which allows it to infer the presence of the step once it bumps into it causing a stepping reflex.
Our task formulation, as well as the use of an asymmetric actor-critic, are the key components to discovering this behavior.
In fact, teacher-student architectures did not enable stepping reflexes, as the learned teacher policies relied heavily on detecting the precise positioning of steps through perception.

\begin{figure*}
  \centering
  \begin{minipage}[t]{0.19\textwidth}
  \centering
    \includegraphics[width = \textwidth, trim={10cm, 10cm, 20cm, 0}, clip]{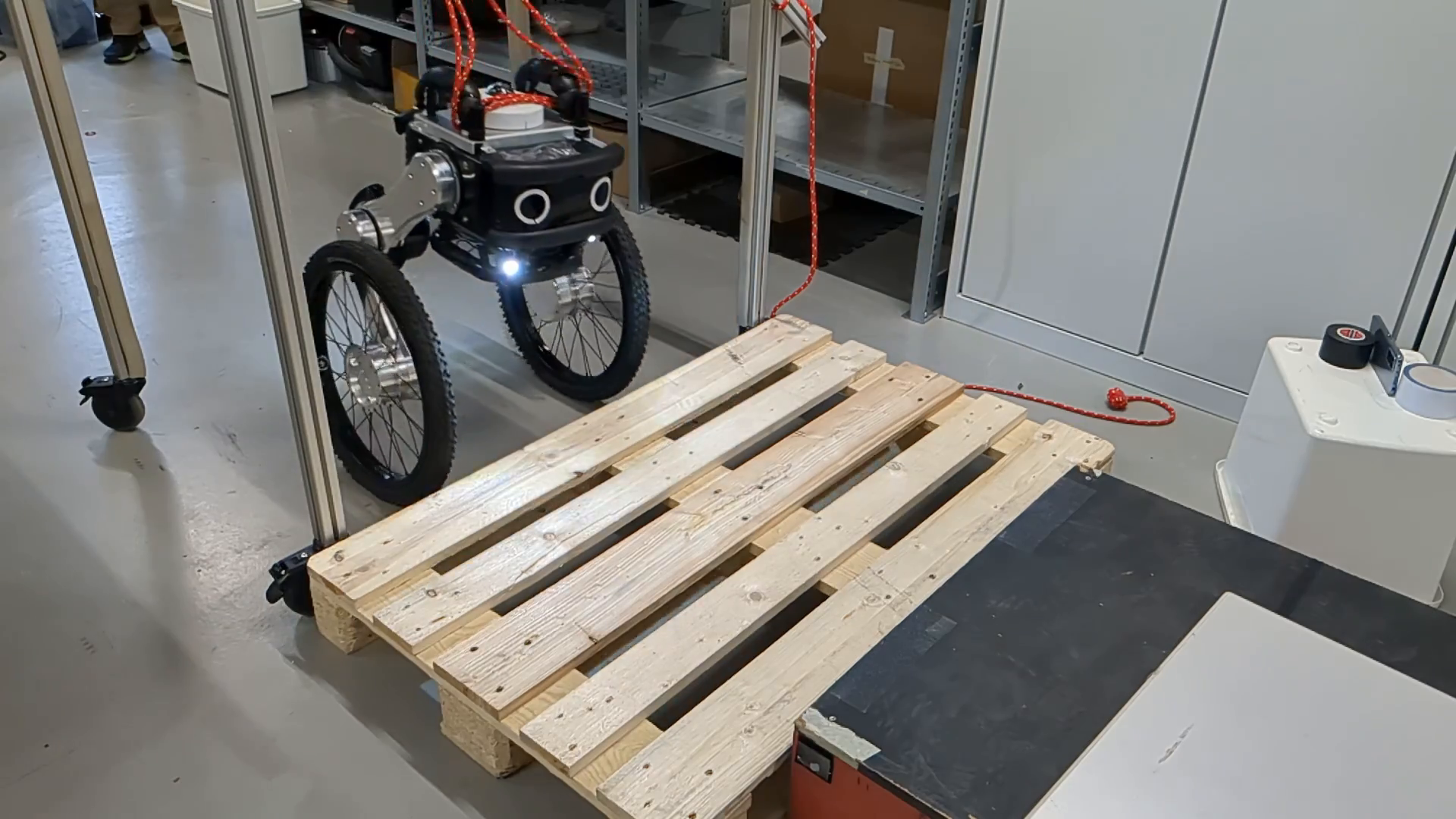}
  \end{minipage}
  \hfill
  \begin{minipage}[t]{0.19\textwidth}
    \centering
    \includegraphics[width = \textwidth, trim={10cm, 10cm, 20cm, 0}, clip]{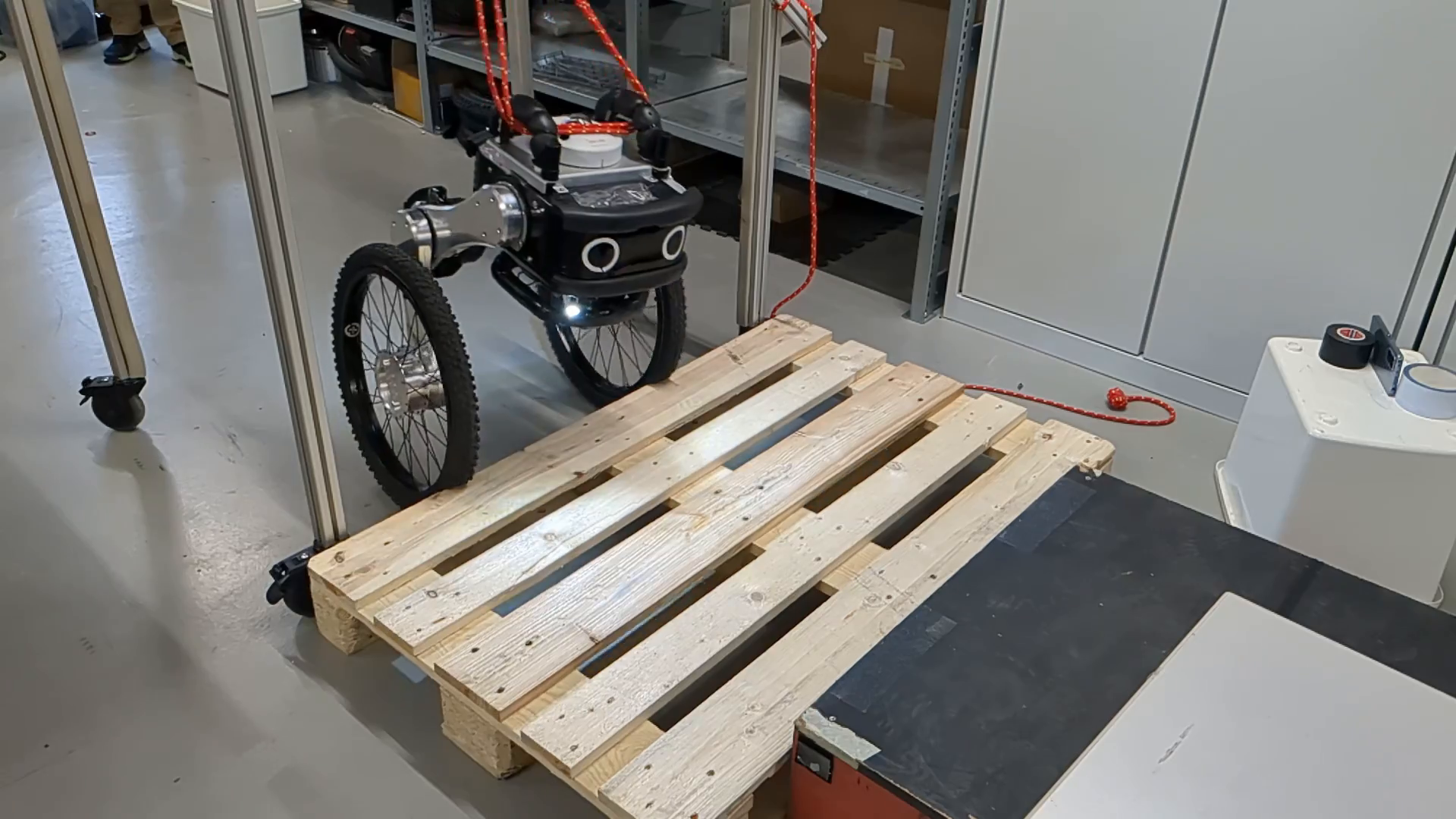}
  \end{minipage}
  \hfill
  \begin{minipage}[t]{0.19\textwidth}
    \centering
    \includegraphics[width = \textwidth, trim={10cm, 10cm, 20cm, 0}, clip]{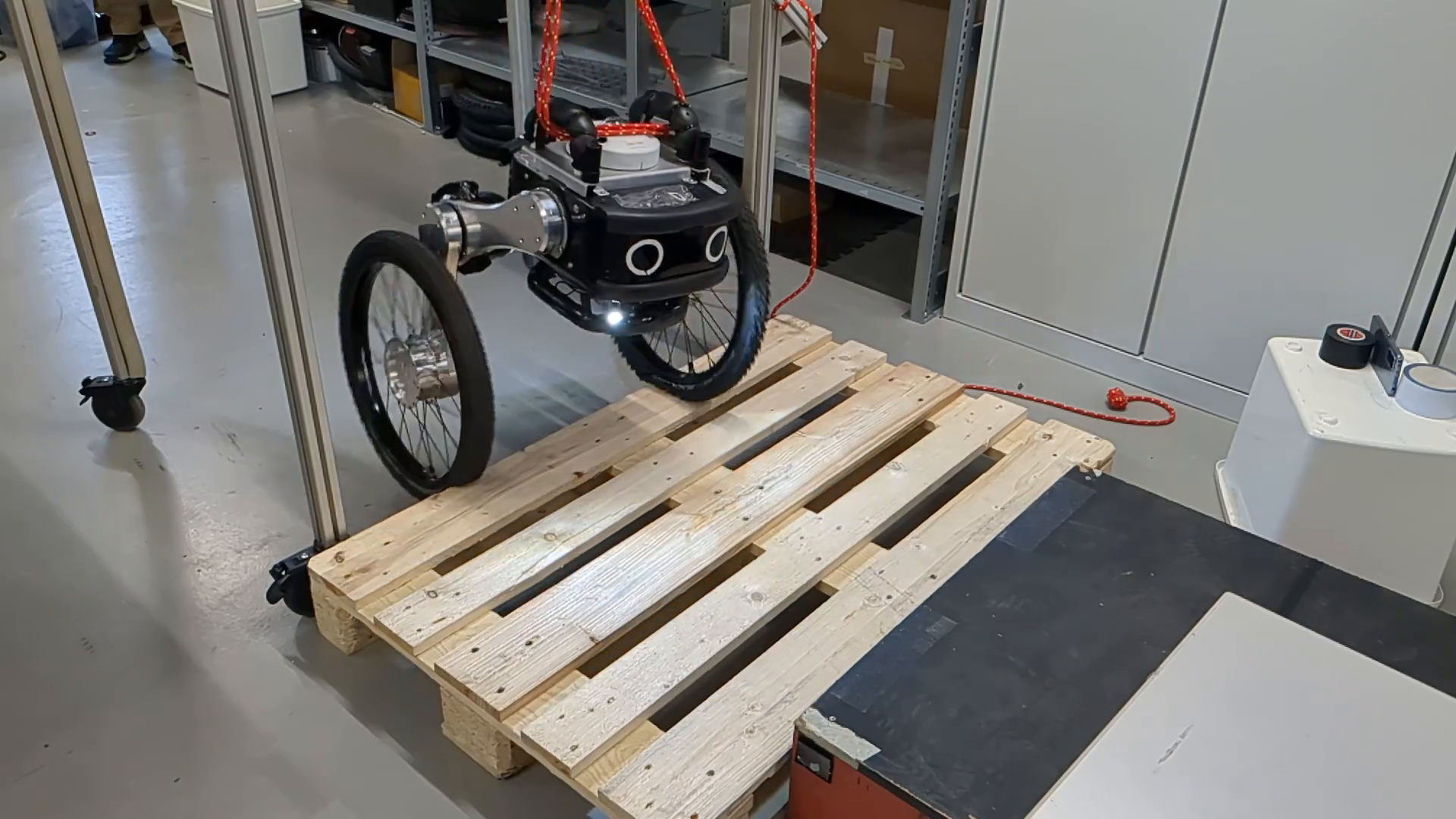}
  \end{minipage}
  \hfill
  \begin{minipage}[t]{0.19\textwidth}
    \centering
    \includegraphics[width = \textwidth, trim={10cm, 10cm, 20cm, 0}, clip]{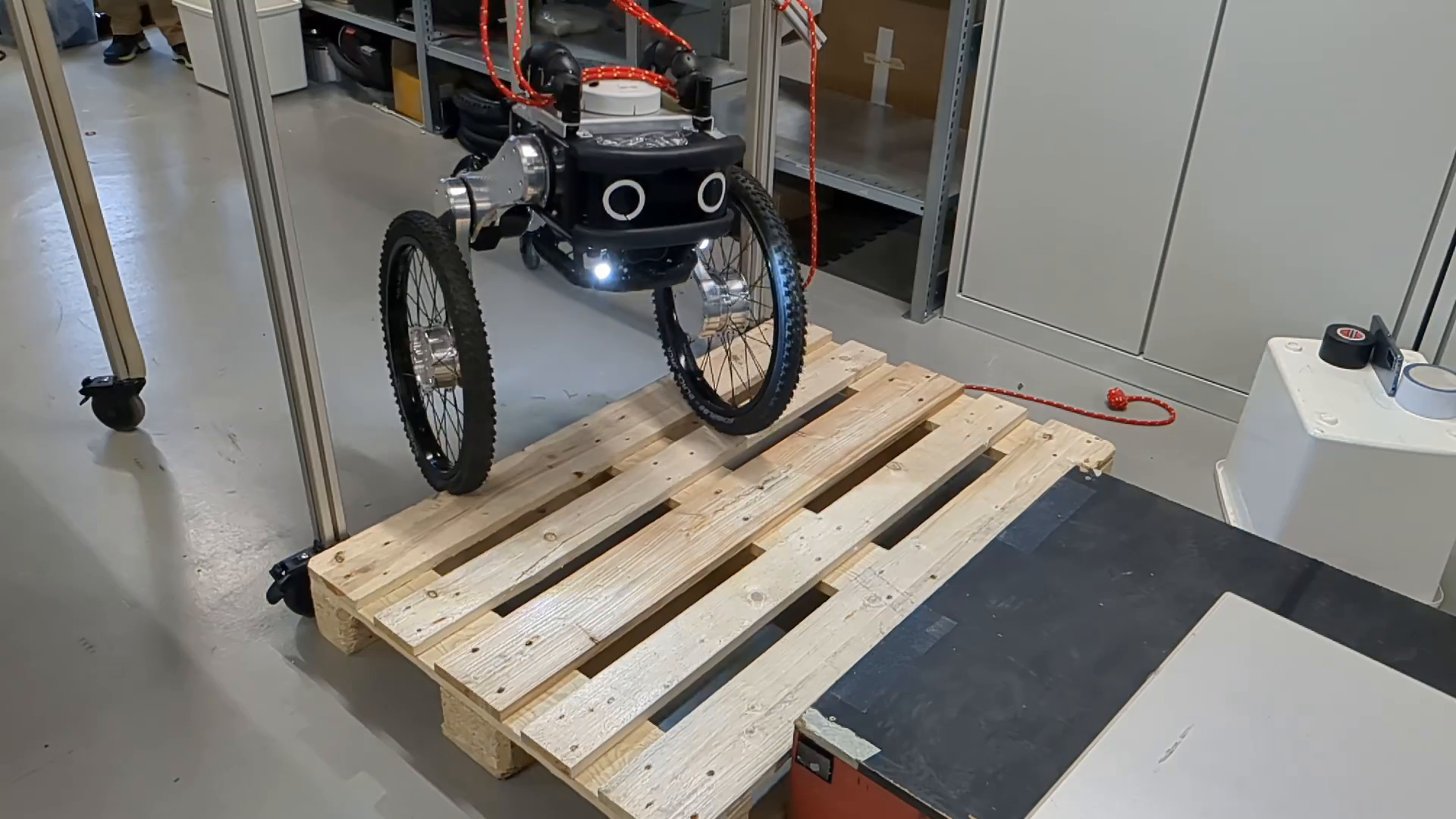}
  \end{minipage}
  \hfill
  \begin{minipage}[t]{0.19\textwidth}
    \centering
    \includegraphics[width = \textwidth, trim={10cm, 10cm, 20cm, 0}, clip]{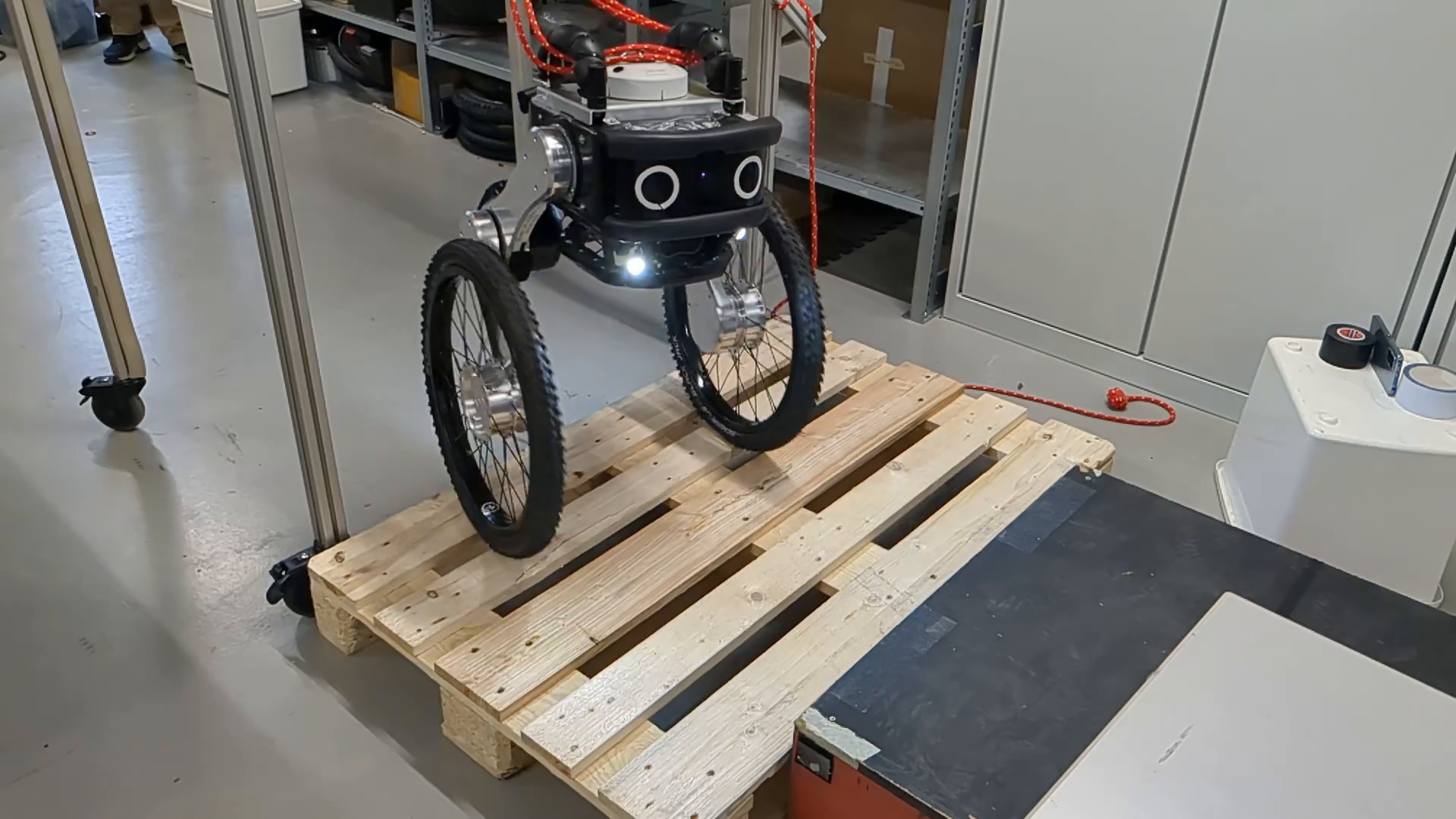}
  \end{minipage}
  \caption{\textbf{Real World Results:} The Ascento robot going up a 15cm step with our learned policy. Images are taken at \SI{4}{Hz}.\vspace{-8pt}}
  \label{fig:robot-results}
\end{figure*}

\subsubsection{Terrain Boolean Importance}
\label{sec:bool}
The importance of the terrain boolean observation for the proposed method can be especially seen when applied to bipeds and wheeled-legged robots.
In fact, it allows the robots to learn specialized knowledge for stair-climbing skills without compromising performance on rough terrain.
Table \ref{tab:results} shows that Cassie, Ascento, and ANYmal on Wheels achieve a higher success rate when on stairs in the \textit{Bool On} experiment than the \textit{No bool} ablation, whereas the \textit{Bool Off variant} outperforms it in other terrains.
The same policy successfully learns these two behaviors conditioned on the boolean observation.
For example, we notice that our learned policy for Cassie adopts a slower gait with higher strides when the terrain bool is present. 
It allows it to maintain more stability and climb over the steps when its feet collide with it. 
When we turn the bool off, it adopts a quicker gait which is more efficient in the other terrains.
Similarly, Ascento showcases a bigger leg split when in stair mode, a position allowing it to lift its front leg when detecting a step through proprioception.
When omitting the terrain boolean in the \textit{No bool} ablation, a compromise has to be made to maximize performance on all terrains, which results in a less specialized policy.
On the other hand, training completely separate policies would result in less robustness as they would not benefit from all the simulated experiences. 
Finally, the ANYmal on Wheels also exhibits interesting behavior. 
In both quadrupedal and bipedal modes, it learns to mostly use its wheels to locomote and step when encountering a step. 
This behavior is not developed without the boolean, hence the low success rates. 

\subsection{Real World Results}
\label{sec:real-world-results}
Our policy was able to climb 15cm steps with Ascento.
Figure \ref{fig:robot-results} illustrates its behavior step by step, and we also show successful climbs in the supplementary video\footnote{Video link: \href{https://youtu.be/Ec6ar8BVJh4}{https://youtu.be/Ec6ar8BVJh4}}.
We can see that the robot drives toward the step with a slight leg split and some speed.
The left wheel initially makes contact with the step and, with some grip, starts moving up.
There is some slippage between the wheel and the step, but as the robot infers the presence of the step from its proprioceptive state, it lifts the left leg up.
Then, the right leg follows and the robot stabilizes at the top of the step.
These results show that our method is promising for complex tasks such as stair-climbing, as this version of the Ascento robot was previously unable to overcome steps of this height with any controller.
During our experiments, we noticed that it is important to compromise between a behavior that uses velocity and grip with the step, versus a behavior that relies only on stepping up one leg at a time.
We achieve this by tuning the shaping rewards described in section \ref{sec:rewards}.
On one hand, a policy that adopts the formerly described behavior will fail due to slippage during the kick-back.
Colliding with the step causes a momentary loss of contact, therefore making it impossible to go up, as collisions and friction with the edge of the step are not perfectly modeled in the simulation.
On the other hand, a behavior relying only on stepping up one leg at a time fails due to the sim-to-real gap in actuation response for aggressive maneuvers.\footnote{The use of real-world data-based actuator models during training, as e.g. in \cite{hwangbo2019learning}, could further aid sim-to-real transfer in the future.}

\subsection{Sim-to-real Transfer}\label{sec:delays}
As discussed in section \ref{sec:sim-to-real}, when deploying a learned policy in the real world, there are delays that are not captured by the simulation~\cite{8593894}.
This is especially important for the purpose of climbing steps because it is a very dynamic motion where timing is crucial.
In our supplementary material video, we show that policies learned without accounting for this sim-to-real gap are very unstable.
They are not robust to disturbances and have very jerky behavior when going up a step.
When climbing a step with Ascento, the joints exhibit high-frequency oscillations and some joint velocities reach a peak of 15 rad/s.
After adding the delay randomization, we can see that the policies are more robust and smoother, successfully climbing the same step without any joint exceeding a velocity of 1.5 rad/s. 

\section{Conclusion}
\label{chap:conclusion}
In conclusion, we propose a method using \ac{RL} to train a controller that equips legged and wheeled-legged robots with stair-climbing capabilities.
Our proposed methodology diverges from the traditional velocity-based approach, favoring a position-based \ac{RL} task. 
We use an asymmetric actor-critic, leveraging privileged information during training, and removing the need for this data during deployment. 
We introduce a boolean observation that serves as a mode switch for stair-climbing, a feature that proves instrumental in achieving good results at this task while remaining robust in other terrains for bipedal robots.
The application of these techniques and methodologies led to the creation of a ``blind" controller that does not require perceptive sensor information or a positioning system. 
Our method allowed Ascento to climb 15cm steps, an achievement previously impossible with this robot prototype. 

\subsection{Limitations \& Future Work}\label{sec:limitations}
While the implementation and successful real-world application of our \ac{RL} controller provides a strong foundation for further improvements and extensions, some limitations remain. 
First, although our method does not require any precise information about its surrounding terrain, it does need access to the proposed terrain boolean. 
An interesting direction for future work could explore the possibility of inferring this boolean observation from raw perceptive data such as images.
The concept of a terrain boolean could be extended further to include multiple, one-hot-encoded terrains, or even an approximate latent representation. 
This concept relates to Rapid Motor Adaptation, where the policy learns to infer a representation of physical parameters from the robot's state history~\cite{kumar2021rma, fu2023deep}.
Alternatively, raw sensor information such as depth imagery could be added to the observation, similarly to~\cite{cheng2023extreme}. However, this would increase complexity and reliance on perception.






\section*{ACKNOWLEDGMENT}
This work was supported by the Quebec Research Funds (FRQNT), the Institute for Data Valorization (IVADO), and Mitacs through their Globalink program.

\newpage
\balance
\bibliographystyle{IEEEtran.bst} 
\bibliography{IEEEexample} 

\end{document}